\documentclass[letterpaper]{article} % DO NOT CHANGE THIS
\usepackage{aaai2027}  % DO NOT CHANGE THIS
\usepackage[hyphens]{url}  % DO NOT CHANGE THIS
\usepackage{graphicx} % DO NOT CHANGE THIS
\usepackage{natbib}  % DO NOT CHANGE THIS AND DO NOT ADD ANY OPTIONS TO IT
\usepackage{caption} % DO NOT CHANGE THIS AND DO NOT ADD ANY OPTIONS TO IT
\usepackage{algorithm}
\usepackage{algorithmic}
\usepackage{amsmath}
\usepackage{newfloat}
\usepackage{listings}
\DeclareCaptionStyle{ruled}{labelfont=normalfont,labelsep=colon,strut=off} % DO NOT CHANGE THIS
\floatstyle{ruled}
\newfloat{listing}{tb}{lst}{}
\floatname{listing}{Listing}

\usepackage{booktabs}

\title{TRIBE: Predicting Team Performance via Communication Behavior Ensembles}
\author{
    Ali Jalal-Kamali\textsuperscript{\rm 1},
    Nikolos Gurney\textsuperscript{\rm 1},
    David V. Pynadath\textsuperscript{\rm 2},
    Fred Morstatter\textsuperscript{\rm 3}
}
\affiliations{
    \textsuperscript{\rm 1}USC Institute for Creative Technologies\\
    \textsuperscript{\rm 2}Rice University Ken Kennedy Institute\\
    \textsuperscript{\rm 3}USC Information Sciences Institute\\
    
}

\nocopyright
\begin{document}

\maketitle

\begin{abstract}
Designing autonomous agents that effectively assist human teams hinges on understanding team dynamics, often without task specific knowledge. We present TRIBE, a domain independent approach that reveals team behavioral dynamics invisible to traditional performance metrics. We show that communication patterns can categorize teams into performance predictive behavioral tribes, as early as 10\% into the task, enabling timely interventions. We test TRIBE on four diverse datasets and demonstrate that communication patterns predict team performance while the prediction strength varies by the degree a task structure allows for behavioral freedom.
Our temporal analysis reveals that AI agents significantly alter team behavioral trajectories while human advisors align with natural dynamics, and that teams maintain behavioral flexibility throughout collaboration. Further, we compare TRIBE to Llama and optimize the pipeline, achieving significant speedup with performance improvement.
\end{abstract}


%%%%%%%%%%%%%%%%%%%%%%%%%%%%%%%%%%%%%%%%%%%%%%%%%%%%%%%%%%%%%%%%%%%%%%%%

\section{Introduction}
\label{sec:introduction}
Autonomous agents increasingly support human teams in complex, real time tasks through automated assessment and assistance during task performance \cite{sukthankar2007,webber2019,seo2021}. However, their ability to understand the people they help often limits the usefulness of agents. The problem is even harder when interacting with teams, since the agent needs to model the communication and behavioral patterns among them. 

Research in human agent interaction has long recognized the importance and difficulty of maintaining accurate models of individuals in collaboration \cite{albrecht2018}. When multiple people work together, communication becomes an integral part of their behavior, a reality documented in diverse domains \cite{emmitt2006,marlow2017,stempfle2002}. This is advantageous for autonomous agents because communication has proven to be a valuable source of information about the characteristics and processes that underlie team performance \cite{tiferes2018}. For an agent seeking to improve team performance, the question is: can we predict which teams struggle, early enough to intervene effectively?

We present TRIBE (Team based Raw Interactions to Behavioral Ensembles), a pipeline that addresses this challenge. We demonstrate that topic modeling combined with clustering extracts meaningful patterns from team communication that are strongly correlated with performance outcomes, enabling early interventions. Through temporal analysis, we show how pre-existing AI agent interventions impact team behavioral dynamics. Lastly, we validate TRIBE across multiple domains and systematically optimize the pipeline for practical deployment.

\section{Related Work}
\label{sec:related}

Traditional team assessment relies on outcome metrics such as task completion time and success rates, which provide clear signals but only after events conclude, limiting utility in ad hoc, high stakes contexts requiring real time intervention. Recent work has shifted toward process oriented approaches using communication patterns \cite{marlow2017,tiferes2018}. \cite{oneill2022} provide a comprehensive review of the human autonomy teaming literature, highlighting the importance of mutual understanding between humans and AI agents. \cite{zhang2021} explore expectations of AI teammates in human AI collaboration, emphasizing that effective teaming requires agents to adapt to human behavioral patterns rather than imposing rigid protocols. \cite{zijlstra2012setting} identified early interaction patterns in swift starting teams, finding that specific communication patterns emerge very quickly and predict effectiveness in ad hoc teams like aviation crews. More recently, \cite{bisberg2025communication} demonstrated that communication patterns in multiplayer online games achieve meaningful predictive power for team skill levels. Entropy based approaches \cite{engome2023entropy} use sliding window entropy to analyze team coordination dynamics, showing that communication entropy can differentiate stable from unstable team communications.

Successful agent intervention hinges on identifying when and how to intervene in human teamwork. Examples from the literature include: \cite{seo2021} propose AI coaches to infer team mental model alignment in healthcare settings. \cite{myers2018} develop autonomous intelligent agents for team training that must understand team state to provide appropriate guidance. Recent work by \cite{li2023markov} introduces Markovian approaches to modeling human trust and reliance in AI assisted decision making that can capture dynamic human AI interaction patterns.
Further work shows effective human-AI teaming requires behavior-aware design \cite{mahmood2024behavior}, learned interaction rules \cite{mozannar2023effective}, and explanation \cite{schleibaum2024adesse}. These works collectively suggest that agent intervention requires understanding team behavioral states.

The approaches outlined have four key limitations. First, they rely on retrospective analysis where the patterns are identified after performance concludes. This limits their applicability in early real time interventions. Second, they are often domain specific: metrics calibrated for gaming communities do not generalize to planning or reasoning tasks. Third, they identify coarse grained patterns without partitioning teams into behavioral categories with clear performance implications. Fourth, agent monitoring frameworks track agent reliability but cannot show how various agents with different logics affect the team behavioral trajectories.

TRIBE addresses all four limitations mentioned above: it enables early prediction, operates domain independently across diverse collaborative contexts, partitions teams into performance predictive behavioral ensembles (tribes), and directly quantifies intervention impact through temporal analysis, revealing how different agents impact the team dynamics independent of proprietary intervention logic.

\section{Data and Task Environments}
\label{sec:team_data}

We analyze team communication from the publicly available (third party) datasets of ASIST program (Artificial Social Intelligence for Successful Teams) \cite{huang2022}, which developed Minecraft task environments to study human teamwork and AI agent intervention. The data was collected at Arizona State University with full IRB approval.

\subsection{Study 3: Urban Search and Rescue }

ASIST Study 3 \cite{huang2022} involved teams of three participants performing collaborative urban search and rescue missions in a Minecraft based environment \cite{corral2021,freeman2021}. Each team member had a unique role: Engineer, Medic, and Transporter. Teams underwent hands-on training followed by two consecutive 17 minute trials. Each trial began with a mandatory 2 minute planning phase where teams remained at the entrance, followed by 15 minutes of gameplay where communication continued. The score is based on the number of rescues in each trial. 

Study 3 involved multiple research teams developing different AI agent systems with proprietary intervention logic and strategies. The teams were distributed across these AI agents: Univ. of Southern California: USC agent \cite{pynadath2023effectiveness}, CMU Robotics Institute: CMURI agent \cite{sycara2021asist}, Charles River Analytics: CRA agent \cite{cra2024asist}, Smart Information Flow Technologies: SIFT agent \cite{kuter2022sift}, Univ. of Arizona: UAZ agent \cite{pyarelal2023uaz}, MIT: DOLL agent \cite{simondempseyteam}; plus two categories for a Human-Advisor and No-Advisor, for a total of 8 intervention categories. It is worth mentioning that TRIBE has no information about or access to the inner logic of any AI agent, and it only observes the interactions.  

To inform TRIBE, we analyze only the natural language transcripts of team communications, excluding other telemetry data from the testbed logs. 

\section{The TRIBE Pipeline}

\subsection{Preprocessing of Transcripts}

We implemented a comprehensive preprocessing pipeline to prepare transcripts for analysis. We removed all annotations, encodings, and administrator messages, retaining only team communication transcripts. Since each team performed two trials, we split each team file into two separate trials. After removing duplicates, we obtained 222 trials from 111 teams, with trial transcripts with 80 to 370 lines. 

We created Document-Term Matrix (DTM) representations for each trial using the textmineR package \cite{jones2021}. This process involved converting text to lowercase, removing punctuation and numbers, and filtering standard stopwords. We used unigram models throughout our analysis.

\subsection{Topic Modeling}
\label{sec:team_methodology}

From an agent's perspective, determining which teams may need assistance requires understanding how team communication patterns relate to performance. TRIBE leverages topic modeling to extract latent themes from team conversations, then uses clustering to identify behavioral ensembles (tribes) that correlate with performance outcomes.

\subsubsection{Latent Dirichlet Allocation}

After preprocessing, we apply topic modeling to discover the content structure of team communications. Topic modeling is an unsupervised method that extracts latent topics from text, where topics represent recurring themes characterized by word distributions. Latent Dirichlet Allocation (LDA) \footnote{Due to space constraints, for all statistical, AI, and machine learning methods, we provide citations for detailed descriptions.} \cite{blei2003lda} is the standard method for topic modeling and we used the R package textmineR \cite{jones2021}. 

The first step was to determine the optimal number of topics. To do so, we evaluated average probabilistic coherence \cite{mimno2011optimizing}, a metric of topic quality. To ensure stability, we evaluated coherence per topic count (2-20) by averaging 100 runs, yielding k=12 as optimal.

\subsection{Behavioral Clustering}
\label{sec:clustering}

To identify potential behavioral categories among teams that indicate different performance levels, we perform clustering over the topic probability distributions. To do so, each trial is represented by its topic distribution vector (the theta matrix from LDA), which captures the proportional probability of each topic in each trial's transcript.

We used gap statistics \cite{tibshirani2001estimating} with 500 iterations to determine the optimal cluster count, yielding 8 clusters. We then applied k-means clustering to assign each of the 222 trials to one of these 8 clusters, representing distinct behavioral patterns.

\subsection{Performance Prediction}
\label{sec:performance_prediction}

To evaluate whether TRIBE's behavioral clusters relate to trial scores, linear regression revealed a significant relationship between cluster assignment and trial scores, indicating strong correlation between communication patterns captured by clusters and performance outcomes. Table \ref{tab:team_trial_separation} contains the cluster sorted by their mean score, how they are split between first and second trials, and the significance of that split.

\begin{table}[!h]
\centering
\caption{Cluster trial separation, ranked by score mean.}
\label{tab:team_trial_separation}
\setlength{\tabcolsep}{3pt}
\begin{tabular}{c|c|cc|c}
Cluster & Score (Mean$\pm$SD) & 1st Trial & 2nd Trial & $p$ \\
\hline
 1 & 635 $\pm$131 & 14\% & $\boldsymbol{86\%}$ & $<$0.001 \\ 
 2 & 617 $\pm$128 & 29\% & $\boldsymbol{71\%}$ & 0.024 \\
 3 & 614 $\pm$170 & 29\% & $\boldsymbol{71\%}$ & 0.036 \\
 4 & 608 $\pm$149 & 11\% & $\boldsymbol{89\%}$ & 0.001 \\
 5 & 554 $\pm$133 & $\boldsymbol{91\%}$ & 9\% & $<$0.001 \\
 6 & 502 $\pm$147 & $\boldsymbol{84\%}$ & 16\% & $<$0.001 \\
 7 & 470 $\pm$123 & $\boldsymbol{84\%}$ & 16\% & 0.001 \\
 8 & 421 $\pm$116 & 53\% & 47\% & $\boldsymbol{1.00}$ \\
\end{tabular}
\end{table}

As shown in the table, seven of eight clusters show a significant trial skew (binomial $p<$ 0.05), meaning TRIBE differentiates trial order from communications alone, with no knowledge of trial order, semantic content, or scores. This is not simply tracking the score increase between trials since scores alone poorly separate 1st and 2nd trials (pseudo-$R^2$=0.08), while TRIBE clusters do so far more strongly (pseudo-$R^2$=0.31). Only the worst performing cluster shows no skew (53\%, 47\%; $p=$ 1.00) indicating poor team behavior is trial agnostic, which is a strong signal for intervention if we can predict it early on.

\subsection{Early Prediction: Real Time Viability}
\label{sec:early_prediction}

A critical capability for any intervention system is the ability to assess team status early enough to enable timely assistance. To test TRIBE's viability for real time monitoring, we evaluated prediction accuracy when classifying teams based on progressively larger transcript segments.

The methodology: for each trial, we extracted cumulative 10\% increments of the transcript and inferred each segment's topic distribution via fold-in Gibbs sampling against the original model's fixed topic-word distribution, rather than refitting a new topic model per segment. This way we ensured that the segment and the full trial theta vectors share an identical topic space. Each segment was then assigned to its nearest of the 8 original cluster centroids and compared against the trial's full transcript cluster assignment.

\textbf{Results:} TRIBE achieves 47\% prediction accuracy at 10\% of the transcript, a 3.0x improvement over the majority cluster baseline of 15.8\% (and 3.76x over uniform random of 12.5\%), rising to 76\% at 30\%, and 90\% at 50\%. 

These results establish proof of concept that early communication patterns can predict behavioral cluster membership with substantial accuracy. Given that a team starting in a high performing cluster at 10\% may migrate to lower performance by 50\%, such checkpoints represent potential intervention opportunities where an external agent could detect and respond to behavioral degradation. 

\section{Behavioral Dynamics Under Various Intervention Strategies}
\label{sec:pipeline_analysis}

The previous section demonstrated the proof of concept: TRIBE captures meaningful performance clusters and enables early prediction. We now investigate deeper questions: What happens when the agents intervene? Do different agent designs produce different impacts? Can we characterize the temporal dynamics of behavioral change?

To answer these questions, we analyze the (pre-existing) intervention metadata embedded in Study 3: teams were assigned to 8 different intervention conditions, the 6 different AI agents, Human-Advisor, and No-Advisor. We apply temporal analyses namely survival analysis, intervention effectiveness quantification, and Markov chain modeling to reveal how each agent shapes team behavioral trajectory. A critical emphasis of this work is that TRIBE assesses intervention impact regardless of each agent's underlying intervention logic. We do not validate or compare the proprietary mechanisms by which these agents decide to intervene. Rather, we demonstrate that TRIBE can reveal the behavioral consequences of interventions across different agent systems, enabling a domain independent approach to understanding how various AI architectures affect team dynamics.

\subsection{Survival Analysis: Behavioral Stability}
\label{sec:survival_analysis}

Survival analysis \cite{Kalbfleisch2002} provides a framework for understanding team behavioral stability over time. To perform survival analysis, we split trials into 10\% segments and model cluster persistence as a survival problem, where the event is the transition from initial cluster.

Of the 8 intervention categories, No-Advisor, SIFT agent, and CRA agent resulted in 100\% of teams changing clusters, and for the remaining 5, most teams changed cluster in the 2 first segments.
The universal median survival time of 1.0 window revealed that most teams abandon initial communication patterns between 10\% and 20\% marks of trials. This timing aligns with Study 3's task structure: each 17 minute trial begins with a 2 minute planning phase followed by 15 minutes of field execution. The cluster change at 10-20\% indicates that TRIBE accurately captures the shift from planning to action which is the real communication behavior shift rather than just being noise in communication patterns.

\subsection{Intervention Effectiveness Analysis}
\label{sec:intervention_effectiveness}

Beyond understanding when interventions cause teams to change clusters, we evaluate whether these impacts improve or degrade team performance. We classified all cluster transitions into four categories: Positive Change (transition to higher ranked cluster), Negative Change (transition to lower ranked cluster), Good Stable (remaining in the same high performing cluster), and Bad Stable (remaining in the same low performing cluster). We calculate Net Change as the difference between Positive and Negative changes, providing a single metric for intervention impact. Table \ref{tab:intervention_effectiveness} presents the percentage breakdown.

\begin{table}[!h]
\centering
    \caption{Intervention outcome percentages by agent.}
\label{tab:intervention_effectiveness}
\setlength{\tabcolsep}{4pt}
\begin{tabular}{l|cc|ccc}
%\hline
\textbf{Agent} & \textbf{Bad} & \textbf{Good} & \textbf{Neg.} & \textbf{Pos.} & \textbf{Net}  \\
 & \textbf{Stable} & \textbf{Stable} & \textbf{Change} & \textbf{Change} & \textbf{Change}  \\
\hline
USC & 22.5\% & 25.0\% & 20.0\% & 32.5\% & \textbf{12.5} \\
Human & 15.7\% & 28.3\% & 25.2\% & 30.7\% & 5.5 \\
CRA & 15.1\% & 20.2\% & 32.5\% & 32.1\% & -0.4 \\
UAZ & 13.8\% & 28.9\% & 28.0\% & 29.4\% & -1.4 \\
DOLL & 26.5\% & 13.7\% & 30.8\% & 29.1\% & -1.7 \\
CMURI & 15.2\% & \textbf{37.8}\% & 24.4\% & 22.6\% & -1.8 \\
SIFT & 13.1\% & 18.8\% & 35.0\% & 33.1\% & -1.9 \\
%\hline
\end{tabular}
\end{table}

TRIBE exposes two distinctive patterns in interventions:

\begin{itemize}
    \item Although \textbf{CMURI} agent interventions have the largest impact (37.8\%) keeping good teams stable in their cluster, when CMURI interventions do change team behaviors, they have a negative net effect (-1.8\%), suggesting successful teams may dismiss interventions.
    \item \textbf{USC} agent interventions, while less frequent (80 total vs. CMURI's 217), demonstrate  positive impact when changing team behavior (12.5\% net positive), more than twice the positive impact of Human-Advisor (5.5\%).
\end{itemize}

\begin{figure}[!h]
\centering
\includegraphics[width=\columnwidth]{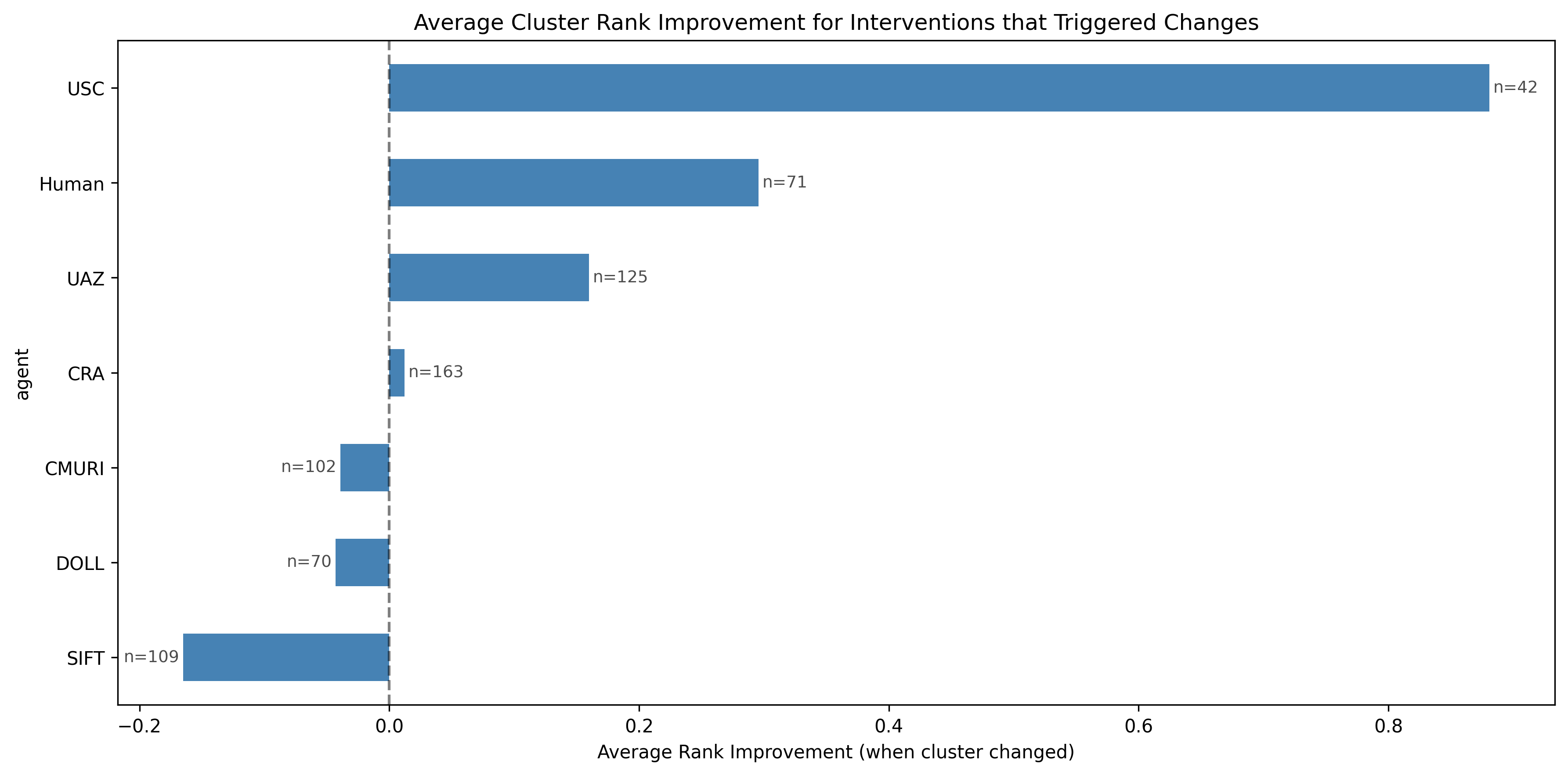}
\caption{Average rank improvement. $n$ is the number of interventions that triggered changes, not the total count.}
\label{fig:rank_improvement_changes}
\end{figure}

To quantify intervention impact depth, we examined cluster rank change post intervention. While two interventions could both have positive effects, one may improve cluster rank more substantially.
Figure \ref{fig:rank_improvement_changes} shows that USC rank improvement is even more impactful, 3x the Human-Advisor impact and more than 4x the UAZ agent. 

To determine statistical significance of observed differences, we compared each agent's transition patterns to No-Advisor baseline using chi square tests, see Table \ref{tab:chi_square}.

\begin{table}[!h]
\centering
\caption{Statistical comparison to No-Advisor baseline. Tests compare
flattened $8{\times}8$ transition matrices. df ranges 53--59 among
tests with exact $p$, varying because cluster-pair cells unobserved
in both conditions are excluded.}
\label{tab:chi_square}

\begin{tabular}{c|c|c|c}
%\hline
\textbf{Agent} & \textbf{$\chi^2$} & \textbf{p-value} & \textbf{Significant} \\
\hline
DOLL & 121.64 & $<$0.0001 & Yes \\
USC & 120.68 & $<$0.0001 & Yes \\
CRA & 118.69 & $<$0.0001 & Yes \\
SIFT & 102.88 & 0.0001 & Yes \\
UAZ & 97.41 & 0.0009 & Yes \\
CMURI & 85.92 & 0.0126 & Yes \\
Human & 57.53 & 0.3113 & No \\
%\hline
\end{tabular}
\end{table}

Notably, Human-Advisor shows no significant difference from No-Advisor ($\chi^2$=57.53, p=0.31), although a larger sample could show a real difference. By contrast, all AI agents show significant divergence (p$<$0.05).

\subsection{Markov Chain Analysis}
\label{sec:markov_analysis}

A Markov chain \cite{norris1998markov} models a system transitioning between discrete states according to fixed probabilities. In our context, states are the 8 communication clusters identified by TRIBE, transitions represent movement between clusters at each 10\% segment, and the Markov property assumes future cluster depends only on current cluster, not history. 

Each agent's influence on team dynamics is represented as an 8×8 transition probability matrix, where P[i,j] is the probability of transitioning from cluster i to cluster j.

Analysis of 252 transitions (28 trials per agent $\times$ 9 transitions) per agent reveals several patterns. Surprisingly, no clusters showed absorption behavior ($>$90\% self transition probability), indicating teams remain behaviorally dynamic throughout trials, no communication pattern becomes stuck, even poor performing clusters. 

\begin{figure}[!h]
\centering
\includegraphics[width=\columnwidth]{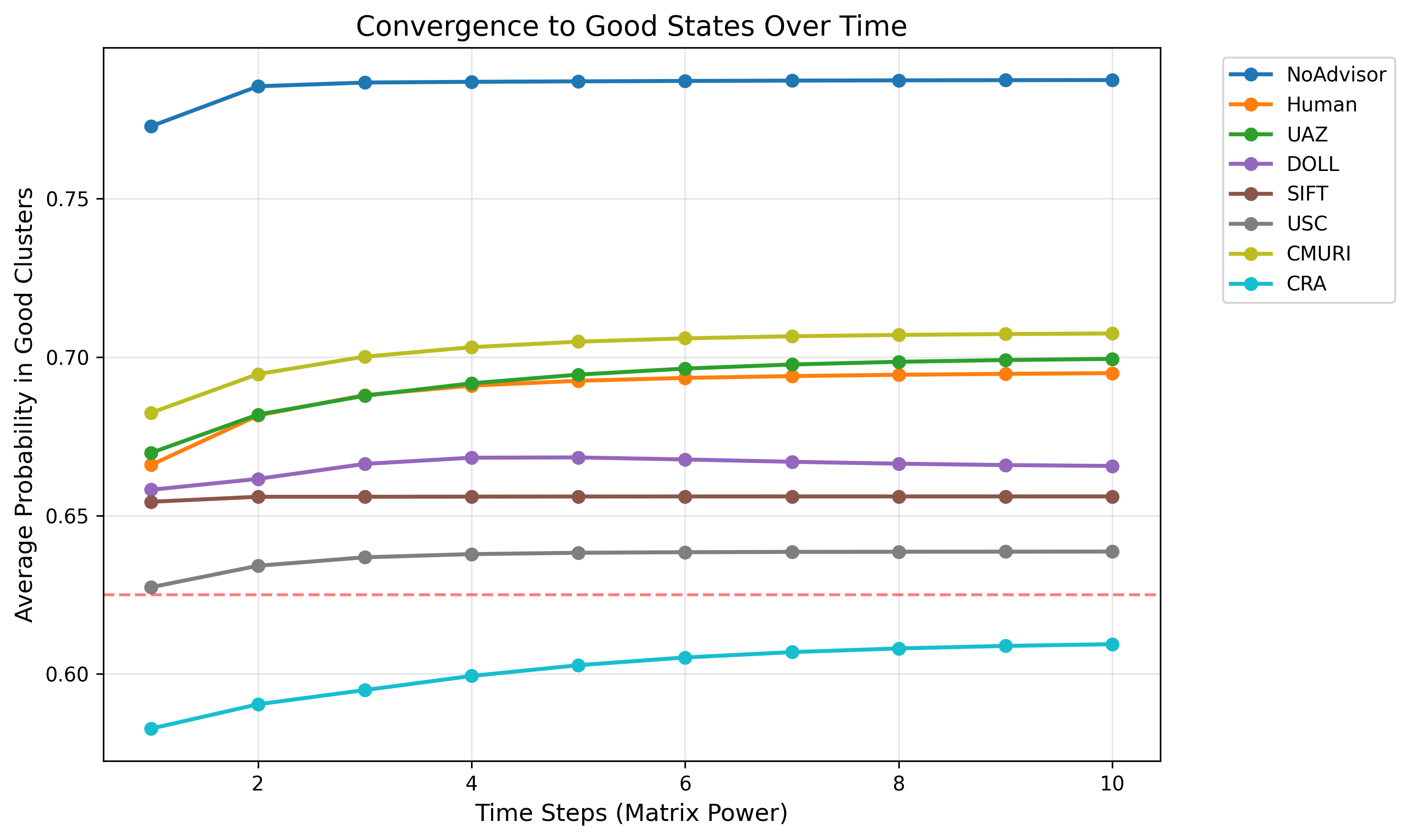}
\caption{Good cluster evolution over 10 time slots.}
\label{fig:evolution}
\end{figure}

Every cluster can reach every other cluster, with varying probabilities, revealing that no behavioral states are isolated and improvement is always possible from any starting point.
Figure \ref{fig:evolution} shows how the probability of being in good clusters (3 best performing clusters: 1, 2, 3) evolves.

No-Advisor starts at 77\% and stabilizes around 78\%, showing natural dynamics favor good clusters. CMURI, UAZ, and Human-Advisor start around 66-67\% and improve to 70\%. USC, SIFT, and DOLL start and remain in 63-66\% range. CRA starts lowest at approximately 58\% with minimal improvement. Most changes occur by step 2-3, confirming that the teams quickly establish behavioral patterns persisting throughout trials.
We calculate Mean First Passage Time (MFPT) \cite{feller1968introduction} to quantify the average steps needed to transition from worst to best cluster; as shown in Table \ref{tab:mfpt}. 
\begin{table}[!h]
\centering
    \caption{Average steps to reach best cluster from worst.}
\label{tab:mfpt}
\begin{tabular}{l|c}
%\hline
\textbf{Agent} & \textbf{MFPT (steps)} \\
\hline
No-Advisor & 8.1 \\
USC & 11.2 \\
Human & 12.7 \\
CMURI & 20.1 \\
%\hline
\end{tabular}
\end{table}

The fact that No-Advisor provides the fastest path from worst to best performance is striking, yet, it does raise questions about the fairness of team quality distribution over the agent cases vs. the No-Advisor case. We checked whether team quality was fairly distributed across the 8 categories. Table  \ref{tab:cluster-agent} shows the distribution of trial quality per category and reveals that No-Advisor received disproportionately high quality teams while USC received low quality teams, explaining why No-Advisor achieved faster MFPT. 

\begin{table}[!h]
\centering
\caption{Trial count by agent, based on cluster ranking.}
\label{tab:cluster-agent}

\setlength{\tabcolsep}{1pt}
\begin{tabular}{c|c|c|c|c|c|c|c|c}
%\hline
Cluster & CMURI & CRA & DOLL & SIFT & UAZ & USC & Human & No-Adv \\
\hline
1 & 5 & 5 & 5 & 2 & 2 & 5 & 4 & 7 \\
2 & 4 & 6 & 2 & 9 & 3 & 3 & 4 & 3 \\
3 & 4 & 4 & 5 & 1 & 6 & 1 & 1 & 6 \\
4 & 3 & 1 & 0 & 3 & 3 & 2 & 3 & 3 \\
5 & 5 & 1 & 6 & 2 & 4 & 4 & 6 & 4 \\
6 & 0 & 6 & 5 & 8 & 3 & 4 & 3 & 2 \\
7 & 5 & 3 & 1 & 2 & 2 & 5 & 4 & 3 \\
8 & 2 & 2 & 4 & 1 & 5 & 4 & 1 & 0 \\
%\hline
\end{tabular}
\end{table}

No-Advisor has 0 trials in worst performing cluster and largest number in best cluster. Aggregating clusters into three performance tiers (top: 1,2,3; middle: 4,5; low: 6,7,8) we arrive at the distributions in Table \ref{tab:performance-groups} which indicate that No-Advisor and CMURI were assigned disproportionately high quality teams, whereas USC received predominantly low performing teams. Such allocation bias in Study 3 dataset, rather than inherent differences in natural team dynamics, can explain the counterintuitive finding that unguided teams outperformed those with AI assistance in MFPT.

\begin{table}[!h]
\centering
\caption{Trial count by agent and performance group.}
\label{tab:performance-groups}
\setlength{\tabcolsep}{1pt}
\begin{tabular}{c|c|c|c|c|c|c|c|c}
%\hline
Tier & CMURI & CRA & DOLL & SIFT & UAZ & USC & Human & No-Adv \\
\hline
Top & 13 & 15 & 12 & 12 & 11 & 9 & 9 & 16 \\
Mid & 8 & 2 & 6 & 5 & 7 & 6 & 9 & 7 \\
Low & 7 & 11 & 10 & 11 & 10 & 13 & 8 & 5 \\
%\hline
\end{tabular}
\end{table}

Starting tier distributions differ across agents, so cross agent comparisons should be read jointly with the rank improvement analysis. Critically, because clusters derive purely from communication and are independent of trial score, detecting this imbalance by score alone would be circular; meaning labeling high scoring teams as better and citing those same scores as evidence of bias explains nothing about \emph{why} they scored well. Was it that stronger teams were assigned to No-Advisor; or was it that withholding intervention improving performance. TRIBE resolves this: the imbalance is attributable to team composition, not intervention effect.

\section{Method Comparison and Optimization}
\label{sec:method_optimization}

Having established that TRIBE works for performance prediction and intervention analysis, we now ask two complementary questions: 
(1) Does a general purpose LLM, deployable locally within an agent, extract performance-predictive behavioral patterns without task specific fitting? We test whether a modern Large Language Model with vast semantic knowledge can extract meaningful behavioral patterns from team communication, potentially improving on our approach.
(2) Can we optimize the pipeline beyond the baseline? The baseline TRIBE uses LDA for topic modeling and k-means for clustering. We systematically test alternative methods for each component to determine whether different combinations might achieve better performance prediction.
(All temporal analysis in previous section employed the LDA+k-means configuration to ensure fair intervention assessment.)

\subsection{Llama Behavioral Extraction}

We tested whether Llama 3.3 70B via Ollama \cite{dubey2024llama}, a modern LLM with extensive training on collaborative task data, could extract behavioral patterns from team communication. At Stage 0, we provided the LLM with the dataset, without any mention of clusters and asked for behavioral patterns that could predict the performance, but the outcomes were uninformative. Then, we proceeded in three stages with progressively more specific prompts.

 \textbf{Stage 1 - Free labeling prompt}: We let the LLM label freely with the prompt: \textit{"Based on team communication patterns, create a 2 word label for each cluster that reflects its behavioral patterns."}
The labels were too generic and all clusters received positive labels regardless of performance.
 
 \textbf{Stage 2 - Domain informed prompt}:  We gave Llama domain information and made it clear that we were looking for labels based on behaviors and performances, asking for differentiation and uniqueness.
The labels were still extremely generic and not indicating anything about performance as the worst cluster was labeled "Good Communication."

 \textbf{Stage 3 - Performance informed prompt}:  We provided all detailed information about the task and data, crucially provided the \textit{cluster score means} along with \textit{cluster performance rankings}, and asked for behavior and performance based labeling.
 Only the explicit ranking of the cluster made the LLM come
up with somewhat relevant labels, and yet, Cluster 4 was labeled "Rank-4 Team" (Table \ref{tab:llama}), meaning it failed to extract meaningful behavioral information.

\begin{table}[!h]
\centering
\caption{Llama behavioral labeling based on cluster ranking.}
\begin{scriptsize}
\setlength{\tabcolsep}{2pt}
\label{tab:llama}
\begin{tabular}{c|l|l|l}
%\hline
Cl. & 1- Free Labeling & 2- Domain informed & 3- Performance informed \\
\hline
1 & Specialized Coordination & Strong Collaboration & Strategic Execution \\
2 & Highly Coordinated & High Quality & Effective Coordination \\
3 & Efficient Rescue & Outstanding Performance & Focused Approach \\
4 & Dynamic Teamwork & Exceptional Coordination & Rank-4 Team \\
5 & Strategic Prioritization & Above Average & Limited Progress \\
6 & Role Based Comm. & Excellent Teamwork & Inefficient Response \\
7 & Real time Collaboration & Very Effective & Disorganized Rescue \\
8 & Task Oriented Dialogue & Good Communication & Chaotic Efforts \\
%\hline
\end{tabular}
\end{scriptsize}
\end{table}

Even with explicit performance rankings and detailed task information, Llama did not identify meaningful behavioral patterns, despite its vast training on collaborative language. This comparison is deliberately scoped as one open weight model under three prompting regimes. Frontier models, fine tuning, or embedding based pipelines may perform better. Our claim is therefore narrower than "TRIBE outperforms LLMs": for real time, on device deployment within agents that must analyze team communications and deliver timely interventions, our lightweight statistical pipeline extracted performance-predictive structure that this LLM configuration did not, despite its vastly greater scale and compute.

\subsection{Topic Modeling Alternative Methods}
\label{sec:topic_modeling_alternatives}

LDA's main limitation is the computational cost ( $>$ 2 hours for 100 runs with reproducibility challenges). We evaluated NMF \cite{lee1999learning,lee2000algorithms}, Top2Vec \cite{angelov2020top2vec}, DTM \cite{blei2006dynamic}, and STM \cite{roberts2014structural} as alternatives. (LSA/LSI lacks topic structure; HDP over fragments small corpora; BERTopic's density based clustering is unreliable below 1k documents.)
NMF emerged as the most promising alternative. With optimal configuration (6 topics), NMF finished in just 5 minutes achieving a 24x speedup over LDA which took 120 minutes. 
Top2Vec consistently discovered only 2-3 topics (insufficient for communication granularity). DTM and STM showed promise theoretically but didn't improve the performance prediction.

Unigram based NMF with k=6 topics, also outperformed bigram variants, becoming optimal topic modeling method for the combined pipeline optimization.

\subsection{Clustering Alternative Methods}
\label{sec:clustering_comparison}

All clustering methods in this comparison operate on the NMF (k=6) topic-distribution vectors identified as optimal in the previous subsection.
Beyond k-means, we tested: Gaussian Mixture Models (GMM) \cite{reynolds2015gaussian} with 4 covariance types, Bayesian GMM (BGMM) \cite{bishop2006pattern}, hierarchical clustering (4 linkage methods) \cite{ward1963hierarchical}, and Dynamic Time Warping (DTW)\cite{sakoe1978dynamic}.

Internal vs. external validation disconnect: Standard clustering metrics (BIC, silhouette, gap statistics) suggested k=3-4 as optimal cluster counts. However, external validation against team performance revealed the opposite that more granular clustering (k=9-16) actually predicted performance better (Table \ref{tab:external_validation_summary}). 
Since $R^2$ increases with the number of clusters and each additional cluster adds a regression degree of freedom, to have a fair comparison across configurations with different k, we report $\bar{R}^2$ (adjusted $R^2$) instead of $R^2$.

To ensure reported values reflect stable estimates rather than a single favorable initialization, each stochastic clustering method (K-means, GMM, BGMM) was refit across 100 random seeds, with hyperparameters selected per fit via unsupervised criteria (BIC or Calinski-Harabasz score), not against trial score. We report the mean $\bar{R}^2$ across 100 seeds.

\begin{table}[!h]
\centering
\caption{Clustering method comparison.}
\label{tab:external_validation_summary}
\begin{tabular}{l|c|c}
Method & k & $\bar{R}^2$ \\
\hline
K-means & 12 & 21.8\% \\
GMM & 16 & \textbf{25.4\%} \\
BGMM & 10 & 20.1\% \\
Hierarchical & 3 & 18.3\% \\
DTW & 6-20 & 14.7–15.1\% \\
\end{tabular}
\end{table}

GMM with k=16 and BGMM with k=9 were the top performers. GMM(16) achieved 25.4\% $\bar{R}^2$ and became our optimal clustering component. 
To confirm the clustering–performance relationship is not an artifact of chance, we ran a permutation test at our selected configuration (GMM, k=16): shuffling trial scores 1,000 times and refitting on each which yielded a $\bar{R}^2$ that never exceeded 0.152 (mean 0.028), far below the observed 0.254 ($p<0.001$).

\subsection{The Optimal Pipeline}
The first thing we tested, before the baseline TRIBE (LDA 12+K-means 8), was TF-IDF (SVD-reduced to 12 dims, k=8) which produced $\bar{R}^2$=0.067. The baseline TRIBE achieved $\bar{R}^2$= 0.185, confirming LDA's topic structure captures signals beyond raw lexical features.
Our optimization analysis showed that NMF achieves the best performance amongst topic modeling methods and GMM amongst clustering methods, meaning NMF(6)+GMM(16) is the overall optimal TRIBE pipeline with  $\bar{R}^2$=0.254. Comparing the baseline TRIBE to the optimal TRIBE, we see a 37.3\% improvement in performance, along with 24x speedup (120m vs. 5m).

\section{Domain Independence}
\label{sec:domain_independence}

A critical test for TRIBE is whether insights generalize beyond Study 3. This section examines domain independence by applying TRIBE to four diverse datasets. To ensure fair cross domain comparison, all analyses in this section use the LDA+K-means pipeline. 

\subsection{Cross Domain Methodology}
\label{sec:cross_domain_method}

Each dataset underwent identical processing to ensure fair comparison. The standardized pipeline involved: preprocessing data by cleaning utterances, setting minimum document length thresholds, and matching team data across trials; then performing topic model evaluation with topic counts ranging from 2 to 25 to find optimal configurations and similarly for clustering configurations. 

We selected 4 datasets representing various collaborative tasks with team conversations and performance measures.

\textbf{Study 3} \cite{ASU/QDQ4MH_2022}: detailed in Data section. 

\textbf{Study 4} \cite{ASU/ZO6XVR_2024}: involves teams of 3 for a virtual search and defusal task in an environment where the team score goes down if an explosive explodes. Unlike Study 3, teams in Study 4 can only communicate during planning phases when gathered in a shop. Once they enter the field to defuse explosives, verbal communication is disabled. Teams can return to the shop multiple times to replan as needed, allowing for extended discussions. We chose minimum of 40 lines per trial, leading to 231 trials from 30 teams. 

\textbf{DELI} dataset \cite{karadzhov2023delidata}:  involves teams solving logical reasoning tasks (Wason selection task) through discussion. In this task design, there is one correct answer that can be logically deduced. Teams of 3-5 people deliberate to identify which cards to check to test logical rules. Dataset contains 333 team deliberations.

\textbf{GAP} dataset \cite{braley2018gap}: involves teams ranking 15 survival items by importance after a hypothetical plane crash. The items are then ranked by domain experts for score allocation. The dataset contains 28 team discussions.

\subsection{Performance Prediction}
\label{sec:cross_domain_clustering}

The key test of TRIBE is how it performs across these 4 different domains. Table \ref{tab:performance_stratification} presents the results:

\begin{table}[h]
\centering
\setlength{\tabcolsep}{2pt}
    \caption{Statistical performance of clustering.  }
\label{tab:performance_stratification}
\begin{tabular}{l|c|c|c|c|c}
%\hline
Domain & ANOVA F & p-value & $\bar{R}^2$ & Best-Worst & Cohen's d \\
\hline
Study 4 & 11.95 & $<$0.001 & 44.3\% & 610 pts & 1.82 \\
Study 3 & 8.27 & $<$0.001 & 18.5\% & 214 pts & 0.94 \\
DELI & 4.10 & $<$0.001 & 14.8\% & 28.3\% & 0.51 \\
GAP & 1.42 & 0.253 & 28.6\% & 23 pts & 0.68 \\
%\hline
\end{tabular}
\end{table}

ANOVA F-tests assess variance across clusters; Cohen's d \cite{cohen1988statistical} quantifies effect size. Study 3 achieved $\bar{R}^2$= 18.5\% with clear cluster differentiation. Cohen's d = 0.94 indicates a large effect size, with 82\% of worst cluster teams scoring below average of best cluster. Study 4's setting yielded more dramatic results with $\bar{R}^2$= 44.3\%. Cohen's d = 1.82 represents a very large effect size, with 97\% of worst cluster teams scoring below average of best cluster.

Study 4's "unlimited" time communication structure allowed for more re-strategizing and behavioral expressions to surface, whereas Study 3's 17 minute time limit for planning and execution combined might have created the urgency for taking quick actions to increase the score, limiting the opportunity for behavioral expression.

DELI achieved a modest $\bar{R}^2$= 14.8\% with limited differentiation. Cohen's d = 0.51 represents a medium effect size, with approximately 69\% of the worst cluster teams scoring below the average of the best cluster. DELI's finite answer structure limits behavioral variation, explaining the low $\bar{R}^2$.

While GAP dataset does allow for more behavioral expressions with $\bar{R}^2$ = 28.6\%, it failed the significance test ($p=0.253$), reflecting small sample inflation. However, this is promising for a similar task with larger data collections. 

The cross domain analysis, not only showcases the domain independence capabilities of TRIBE, but also reveals that the task design determines the capacity for behavioral expressions. Consequently, TRIBE's effectiveness depends on the degree that a task allows for behavioral variation.

\subsection{Principal Component Analysis of Study 4}

To understand Study 4's superior predictive power, we performed PCA on its topic distributions. Study 4's LDA model discovered 13 topics from planning phase communications. Each team's communication pattern can be represented as a 13 dimensional vector showing emphasis on each topic. PCA transforms this 13 dimensional space into a new coordinate system where the first principal component (PC1) captures the maximum variance.

\begin{table}[h]
\centering
\setlength{\tabcolsep}{4pt}
\caption{PC1 vs. Cluster Performance}
\label{tab:performance_rank}
\begin{tabular}{c|c|c|c|c}
Rank & Average Score & SD & Teams & PC1 Position \\
\hline
1 & 941 & 152 & 45 & -1.33 \\
2 & 876 & 228 & 40 & -1.49 \\
3 & 657 & 236 & 30 & -0.53 \\
4 & 629 & 296 & 51 & +0.61 \\
5 & 449 & 249 & 17 & +1.46 \\
6 & 402 & 174 & 48 & +1.64 \\
\end{tabular}
\end{table}

The analysis revealed a striking finding: At the cluster level, PC1 explained $R^2 = 0.927$ of the variance in mean cluster performance ($p = 0.002$) and at the trial level, PC1 explained $R^2 = 0.49$ ($p < 0.001$). This result reveals that Study 4's performance is substantially explained by a single behavioral dimension, where teams can be positioned on a performance continuum of communication patterns.

PC1 represents a weighted combination of all 13 topics, with topics 9 and 7 having strong negative loadings ($-0.427$ and $-0.393$ respectively) and topic 11 having strong positive loading ($+0.481$), at the cluster level. This means teams emphasizing topics 9 and 7 receive negative PC1 values (associated with higher performance), while teams emphasizing topic 11 receive positive PC1 values (associated with lower performance).
In other words, PC1 provides a continuous spectrum from high to low performance communication patterns, enabling real time monitoring for intervention design.

\section{Conclusion}
This work introduced TRIBE, a domain independent approach for understanding and predicting team performance through communication patterns. By applying topic modeling and clustering to team conversations, we demonstrated that behavioral dynamics invisible to traditional performance metrics can be extracted for early intervention. Teams can be classified into performance predictive behavioral clusters as early as 10\% into the task, enabling timely support.

Our analysis revealed several counterintuitive findings. While teams without advisors outperformed those with AI guidance due to bias in team quality assignment, TRIBE's ability to detect this bias validates its capacity to identify meaningful performance related patterns. Human advisors produced patterns statistically indistinguishable from natural team evolution, whereas all AI agents significantly altered behavioral trajectories, showing how TRIBE can evaluate the impact of interventions. TRIBE's survival and Markov analyses revealed behavioral flexibility throughout trials.

Cross domain validation established TRIBE's generalizability, while task characteristics determine how much behavioral patterns emerge. Study 4's planning structure achieved 44.3\% predictive power, with PC1 alone explaining 49\% of trial level performance variance, a continuum enabling real time intervention triggers; whereas DELI's finite answer task reached only 14.8\%.

Method optimization improved predictive power by 37.3\% with a 24x speedup. TRIBE's comparison to a contemporary LLM showed that, despite the LLM's semantic knowledge, it could not identify meaningful behavioral patterns, suggesting the statistical structure of interaction patterns reveals team behavioral signatures. Overall, TRIBE offers a scalable, domain independent approach for team performance prediction where task design allows sufficient behavioral expression, providing actionable insights for autonomous agents supporting teams.

\bibliography{tribe_2027}


\end{document}